\documentclass[10pt,journal,compsoc]{IEEEtran}
\ifCLASSOPTIONcompsoc
  \usepackage[nocompress]{cite}
\else
  \usepackage{cite}
\fi
\ifCLASSINFOpdf
\else
\fi

\usepackage{amsmath,amsfonts}
\usepackage{algorithmic}
\usepackage{algorithm}
\usepackage{array}
\usepackage{textcomp}
\usepackage{stfloats}
\usepackage{url}
\usepackage{verbatim}
\usepackage{graphicx}
\usepackage{cite}
\usepackage{arydshln}
\DeclareMathOperator*{\argmin}{argmin}

\newcommand{\ie}{\textit{i}.\textit{e}.}
\newcommand{\eg}{\textit{e}.\textit{g}.}
\usepackage{makecell}
\usepackage{multirow}
\usepackage{booktabs}
\usepackage{pifont}

\begin{document}
\title{Prompt Tuning of Deep Neural Networks for Speaker-adaptive Visual Speech Recognition}

\author{Minsu Kim,
        Hyung-Il Kim, \IEEEmembership{Member,~IEEE},
        and Yong Man Ro, \IEEEmembership{Senior Member,~IEEE}% <-this % stops a space
\IEEEcompsocitemizethanks{\IEEEcompsocthanksitem M. Kim and Y. M. Ro are with the Image and Video Systems Lab., School of Electrical Engineering, Korea Advanced Institute of Science and Technology (KAIST), 291 Daehak-ro, Yuseong-gu, Daejeon, 34141, Republic of Korea (e-mail: ms.k@kaist.ac.kr; ymro@kaist.ac.kr).
\IEEEcompsocthanksitem H.-I. Kim is with Visual Intelligence Research Section, Artificial Intelligence Research Laboratory, Electronics and Telecommunications Research Institute (ETRI), Daejeon, 34129, Republic of Korea (e-mail: hikim@etri.re.kr). Corresponding author: Y. M. Ro (fax: 82-42-350-5494).}% <-this % stops an unwanted space
}

\IEEEtitleabstractindextext{
\begin{abstract}
Visual Speech Recognition (VSR) aims to infer speech into text depending on lip movements alone. As it focuses on visual information to model the speech, its performance is inherently sensitive to personal lip appearances and movements, and this makes the VSR models show degraded performance when they are applied to unseen speakers. In this paper, to remedy the performance degradation of the VSR model on unseen speakers, we propose prompt tuning methods of Deep Neural Networks (DNNs) for speaker-adaptive VSR. Specifically, motivated by recent advances in Natural Language Processing (NLP), we finetune prompts on adaptation data of target speakers instead of modifying the pre-trained model parameters. Different from the previous prompt tuning methods mainly limited to Transformer variant architecture, we explore different types of prompts, the addition, the padding, and the concatenation form prompts that can be applied to the VSR model which is composed of CNN and Transformer in general. With the proposed prompt tuning, we show that the performance of the pre-trained VSR model on unseen speakers can be largely improved by using a small amount of adaptation data (\eg, less than 5 minutes), even if the pre-trained model is already developed with large speaker variations. Moreover, by analyzing the performance and parameters of different types of prompts, we investigate when the prompt tuning is preferred over the finetuning methods. The effectiveness of the proposed method is evaluated on both word- and sentence-level VSR databases, LRW-ID and GRID.
\end{abstract}

\begin{IEEEkeywords}
Prompt tuning, visual speech recognition, lip reading, speaker adaptation, learnable padding, CNN prompting.
\end{IEEEkeywords}}

\maketitle

\IEEEdisplaynontitleabstractindextext

% \fi
\IEEEpeerreviewmaketitle

\IEEEraisesectionheading{\section{Introduction}\label{sec:introduction}}
\IEEEPARstart{V}{isual} Speech Recognition (VSR) technology \cite{zhou2013compact,saenko2004articulatory,matthews2002extraction} is developed for recognizing speech into text solely depending on visual information (\eg, lip movements) from an input talking face video. It can be regarded as a counterpart of Audio-based automatic Speech Recognition (ASR) that utilizes speech sound as inputs, and is also called lip reading. Even though the technology has drawn big attention \cite{saenko2009multistream,petridis2018end,afouras2018deep,ma2021lira,kim2021cromm,kim2022distinguishing,ma2022visual,shi2022avhubert} with its attractive applications \cite{ke2018computer,cohen2004semisupervised} such as recognizing speech without sound input thus robust to auditory noises and can be utilized in environments that need to be quiet, it is still hard to find that VSR technology is being employed in real-world applications. This is because the VSR system is inherently sensitive to personal lip appearances and movements \cite{assael2016lipnet}, and it shows degraded performances when applied to unseen speakers even if it is developed using a large-scale database with large speaker variations \cite{yang2020SIlipreading,kim2021lip,zhang2021SIlipreading,kim2022udp}. 

In order to mitigate this problem, a speaker-adaptive VSR system can be developed in analogy to speaker adaptation technologies developed for ASR systems \cite{neto1995lin,anastasakos1997mllr,miao2014towardsSAT,klejch2019meta,huang2020SAspeechsynthesis}. Speaker adaptation method is for narrowing the gap between training and testing data distributions by fitting a trained model to unseen test speakers to improve performances during test time. The method attempts to optimize the speech recognition performance by transforming pre-trained models to well operate on one particular speaker or modifying the encoded features of the target speaker to match the pre-trained model, by using a small amount of adaptation data. The most intuitive way is finetuning the pre-trained model on the adaptation data of the target speaker. However, it is not feasible to store and handle each user-specific model since it yields the total number of parameters the same as that of one model multiplicated by the number of speakers. In addition, it requires a relatively large number of adaptation data to achieve optimal performance for the target speaker as the method tunes a large number of parameters. Therefore, an efficient way to adapt the pre-trained VSR model to diverse speakers is necessary, with a small number of parameters and adaptation data.

Recently, Input transformation \cite{elsayed2018AR1,chen2021AR2,dinh2022AR3,zheng2021AR4,neekhara2022AR5,yu2022defending} and Prompting \cite{brown2020gpt3,zhou2022conditional,li2021prompttuning,liu2021prompting,liu2021prompttuning,jia2022vpt,xu2022prompting,smith2022coda,zhou2022learning} have shown the effectiveness of input-level modification of pre-trained models on adapting the models on different tasks or data distributions without modifying the learned weight parameters of pre-trained models. By introducing learnable external parameters to the input of Deep Neural Networks (DNNs), a pre-trained model can perform different tasks from training \cite{elsayed2018AR1,lester2021prompttuning} or be adapted to shifted data distribution \cite{gao2022visual}. Inspired by the recent advances in Input transformation and Prompting, we propose a novel speaker-adaptive VSR framework that utilizes prompt tuning. Specifically, we propose three different types of prompts, addition form, padding form, and concatenation form, which can be jointly utilized for VSR models. Different from the previous prompt tuning methods that are mainly developed with Transformer variant architecture \cite{liu2021prompting,li2021prompttuning,liu2021prompttuning,lester2021prompttuning,jia2022vpt}, the proposed method can also be employed for CNN from the input level to the intermediate layer level.

Moreover, distinct from the previous speaker adaptation methods that modified the extracted feature by introducing additional layers \cite{xue2014fastadaptation,abdel2013codeicassp,miao2015SATivector,abdel2013codecnn}, the proposed prompt tuning does not introduce an additional adaptation network and does not require finetuning of the pre-trained model. It has the advantage of simplifying the adaptation steps so direct adaptation from a pre-trained model is possible, while the previous works need to train an adaptation network after attaching it to the pre-trained model. Finally, the prompt has a much less number of parameters compared to the pre-trained model, so it can be stored and easily handled for each user (\ie, speaker).

Specifically, we propose three different prompts, i) addition form prompt, ii) padding form prompt, and iii) concatenation form prompt. The addition form prompt is the input-level prompt for CNN and has the same shape as the input video frame. It is added to all input video frames consistently to transform the encoded visual feature to well operate on the target speaker. The padding form prompt is the intermediate feature-level prompt for CNN. It operates by replacing the padding of pre-trained CNN that usually has zero, constant, and reflect values. As the padding of CNN is also convolved with convolution kernels, the padding prompt can adapt the encoded visual feature at each intermediate CNN layer to the target speaker. Finally, the concatenation form prompt is similar to Prompting in NLP \cite{li2021prompttuning,liu2021prompting,liu2021prompttuning} and it is concatenated to the input of Transformer in the temporal dimension. We examine the effect of the combinations of the three different prompts and show that by just tuning the prompt, we can improve the performance of the pre-trained VSR model for unseen speakers. Moreover, by analyzing the performance and the number of parameters required for each speaker, we show that the proposed method is preferred over finetuning methods when only a small number of adaptation data is available. We extensively validate the effectiveness of the proposed method in both word- and sentence-level VSR databases, LRW \cite{chung2016lrw} and GRID \cite{cooke2006grid}. Especially, since there is no speaker-annotated VSR dataset obtained in the wild, we annotate the speaker information for LRW \cite{chung2016lrw} dataset and utilize it to confirm the effectiveness of the proposed method in a real-world setting. 

The major contributions of this paper are as follows.
\begin{itemize}
    \item To the best of our knowledge, this is the first work to explore the effectiveness of prompt tuning in VSR. We show that just finetuning prompts can largely improve the performance of VSR model for unseen speakers.
    \item We propose and analyze different types of prompts, the addition prompt, the padding prompt, and the concatenation prompt, that can be jointly utilized for general VSR models which are composed of both CNN and Transformer.
    \item We evaluate the effectiveness of the proposed method with comprehensive experiments on different adaptation data sizes and different prompt types, and we show that the proposed method can even outperform the finetuning method by just introducing about 0.5\% additional parameters compared to the full model.
\end{itemize}

\section{Related Work}
\subsection{Visual Speech Recognition}
Visual Speech Recognition (VSR), also known as lip reading, is the task of predicting speech in the text by watching a silent talking face video \cite{lafon2006data,saenko2005visual,saenko2005production}. Due to the insufficient information and homophene problems, it is regarded as a challenging problem. With the great development of deep learning \cite{hinton2012deep,sainath2015deep}, the performance of VSR has improved significantly \cite{kim2021lip, ma2021end, shi2022avhubert}. 

In word-level VSR, \cite{stafylakis2017resnetlstm} proposed a model architecture with a combination of 3D CNN and 2D CNN for a visual feature encoder and Bi-LSTM as a temporal encoder. Following studies \cite{weng2019twostream, xiao2020deformation} proposed to utilize dynamic information with two-stream CNN \cite{simonyan2014two,ma2019ts,xu2019two} that models the speech from RGB frames and optical flows. \cite{martinez2020mstcn} proposed to model the temporal dynamics using temporal convolutions by proposing Multi-Scale Temporal Convolutional Network (MS-TCN) which has multi-scale temporal receptive fields. \cite{kim2021mmbridge} proposed to use a cross-modal memory network to complement the insufficient information of the visual-only model. In sentence-level VSR, \cite{assael2016lipnet} firstly proposed an end-to-end VSR framework using Connectionist Temporal Classification (CTC) training objective \cite{graves2006ctc} and evaluated their framework on GRID \cite{cooke2006grid} dataset. \cite {chung2017lrs2} proposed a dataset, LRS2, constructed in the wild environment and Sequence-to-Sequence (Seq2Seq) architecture \cite{sutskever2014sequence} which can model the language with its decoder. \cite{afouras2018deep} significantly improved VSR performance by proposing to utilize Transformer \cite{vaswani2017attention} for the temporal encoder (\ie, back-end) and the decoder. Recently, \cite{ma2021end} improved the temporal encoder with Conformer \cite{gulati2020conformer} which is shown improved performance in speech recognition with its local convolution and global self-attention mechanism. \cite{prajwal2022vtp} focused on improving visual front-end architecture by introducing attention so that the model can extract the most salient region from the talking face video.

Some studies have improved VSR performances by focusing on designing training mechanisms for the models \cite{ma2022visual}. \cite{afouras2020asrisallyouneed,zhao2020hearing,ren2021learningfrommaster,ma2021towards} proposed to use knowledge distillation \cite{hinton2015distilling} so the student VSR model can learn from the teacher ASR model or powerful VSR model. \cite{kim2021cromm,kim2022distinguishing} proposed a method that can learn visual-to-audio mapping functions and bring audio representations through the learned mapping using cross-modal memory networks. \cite{chung2016syncnet,ma2021lira,shi2022avhubert} proposed to pre-train the backbone models in a self-supervised way that can utilize large-scale audio-visual databases, and showed its effectiveness with the powerful VSR performances.

Even with the progress in VSR, speaker-adaptive VSR has not well been studied. Since VSR only utilizes visual information, especially lip movements, it is inherently sensitive to personal lip appearances and movements. It makes the pre-trained VSR model show degraded performances when it is applied to unseen speakers that do not appear in the training dataset \cite{assael2016lipnet,kim2022udp}. In this paper, we try to mitigate the speaker-dependency problem of VSR by developing a speaker-adaptive VSR method. Specifically, we introduce prompt tuning methods to VSR, so we can adapt a pre-trained VSR model to the target speaker without modifying the pre-trained model parameters but just tuning the proposed prompts.

%------------------------------------ Figure 1
%#############################################
\begin{figure*}[t]
	\centering
	\centerline{\includegraphics[width=18cm]{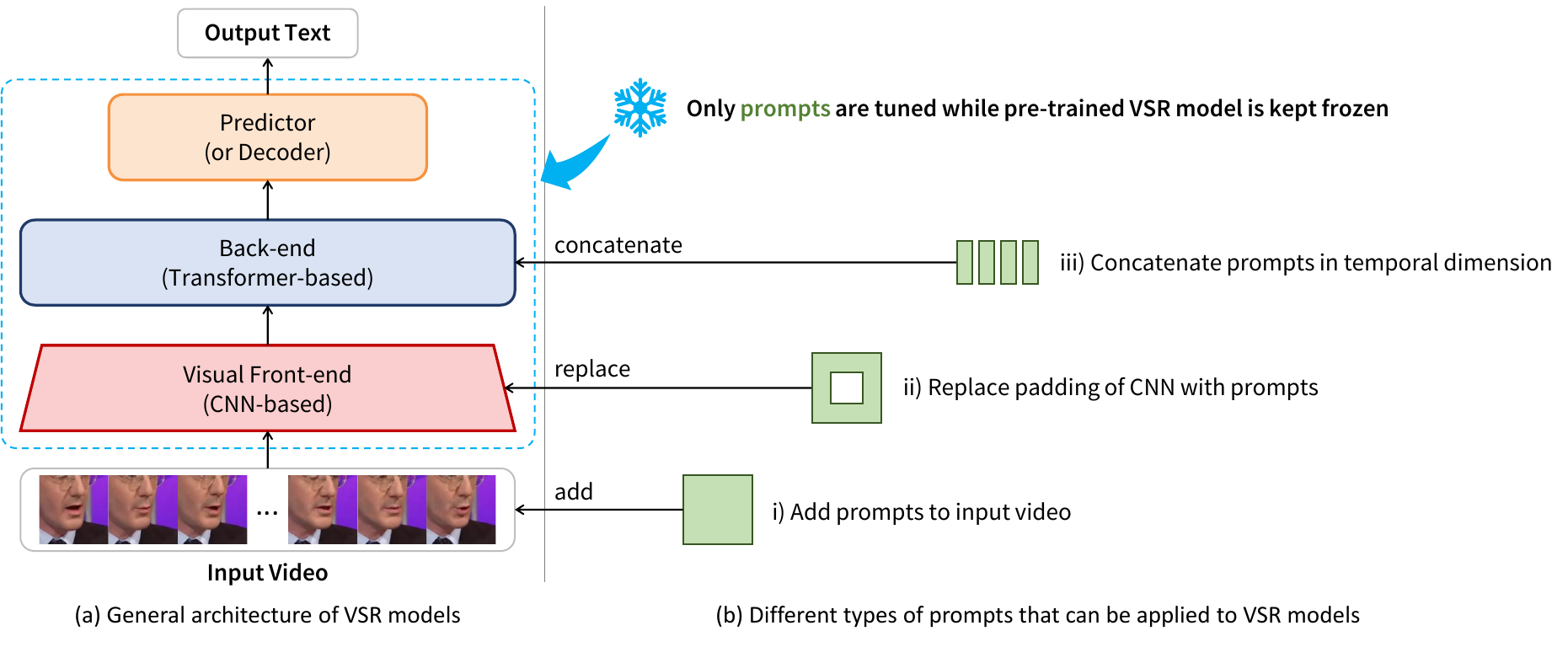}}
	\caption{Illustration of the proposed prompt tuning for speaker-adaptive VSR. (a) The general architecture of VSR models. (b) Three different types of prompts that can be applied to VSR models: i) addition form, ii) padding form, and iii) concatenation form. They can be jointly utilized to adapt the pre-trained VSR model on the unseen target speaker. We only update the prompts while the pre-trained VSR model is kept frozen.
	}
	\label{fig:1}
\end{figure*}
%#############################################

\subsection{Speaker Adaptation}
Speaker adaptation \cite{miao2015SATivector} is mainly developed for ASR. \cite{liao2013trainpart1} propose to finetune different parts of the network to adapt the model to the target speakers. However, as the finetuning methods modify a large number of parameters, they can be suffered from overfitting problems when the adaptation data is not enough. In order to mitigate the overfitting problem, \cite{yu2013kldregularized} proposed a regularization method for speaker adaptation. Some works \cite{seide2011addlayer1, li2010addlayer2} proposed augmenting the pre-trained speech recognition model with additional speaker-dependent layers. \cite{swietojanski2014lhuc} proposed to add a speaker-dependent vector to every pre-trained hidden layer which will be adapted to the test speaker. 
\cite{abdel2013codecnn,abdel2013codeicassp,xue2014fastadaptation} proposed to use speaker codes which are additional inputs depending on speakers. They firstly train an adaptation network by attaching it to the pre-trained model on the train set, and then the speaker codes are optimized for each target speaker on the adaptation set. Therefore, the methods require speaker annotations for both training and adaptation data.
Recently, meta-learning \cite{klejch2019meta} and speech synthesis \cite{huang2020SAspeechsynthesis} based speaker adaptation methods were explored.

A few works handled speaker adaptation for VSR. \cite{almajai2016lipreadingmllt} proposed to utilize traditional speaker adaptation methods of ASR \cite{gopinath1998mllt,anastasakos1997mllr} into VSR. \cite{kandala2019SAlipreading} proposed to bring the concept of i-vector \cite{dehak2010ivector} in VSR. \cite{kim2022udp} proposed to use user-dependent padding for each speaker. In this paper, we propose a novel speaker-adaptive VSR method that utilizes prompt tuning. Different from the previous methods, the proposed method does not need any additional network and has simpler adaptation steps. Moreover, the proposed method requires the speaker information for the target speaker only.

\subsection{Input Transformation}
Recently, input transformation methods attract large attention \cite{chen2021AR2, dinh2022AR3,zheng2021AR4,neekhara2022AR5} with its potential to reprogram a pre-trained model to perform different tasks from training. By surrounding the input image with noises, \cite{elsayed2018AR1} showed that the model originally trained to classify ImageNet \cite{deng2009imagenet} classes can perform hand-written digit classification. Recently, \cite{yu2022defending} showed that using the input transformation, defense against the adversarial perturbs can be achieved. In this paper, we utilize the input transformation concept to develop a speaker-adaptive VSR model, we add learnable parameters to the input video frames for transforming the encoded visual feature to represent well for the unseen target speaker.

\subsection{Prompt Tuning}
Prompting is firstly introduced by \cite{brown2020gpt3} to prepend text instruction to the input to make a pre-trained language model understand a given task. By the prepended text, prompt, the performance of a pre-trained language model can be varied \cite{li2022bridge} and previous works \cite{jiang2020can,shin2020autoprompt,radford2021learning} tried to find the better prompt formula. Recent works \cite{liu2021prompting,li2021prompttuning,liu2021prompttuning,lester2021prompttuning,zhou2022conditional,zhou2022learning} proposed learning the prompt by gradient descent which is named prompt tuning. The prompt tuning is developed mainly for Transformer-based language models and has much fewer parameters compared to the full model parameters but can achieve strong performances. Beyond the language model, the prompt is started to be utilized in visual applications \cite{jia2022vpt,bahng2022visual}. However, the visual prompt was limitedly explored for Visual Transformer (ViT) architecture \cite{dosovitskiy2020vit} and input of CNN \cite{bahng2022visual}.

In this paper, we utilize prompt tuning to adapt a pre-trained VSR model to the target unseen speaker. Different from the previous works, we propose three different types of prompts, addition form, padding form, and concatenation form, which can be jointly utilized for DNNs composed of CNN and Transformer variant architecture. Especially, we introduce intermediate feature-level prompts for CNN through a padding form prompt and show its effectiveness becomes larger according to the model size of CNN.

%------------------------------------ Figure 2
%#############################################
\begin{figure*}[t]
	\centering
	\centerline{\includegraphics[width=19cm]{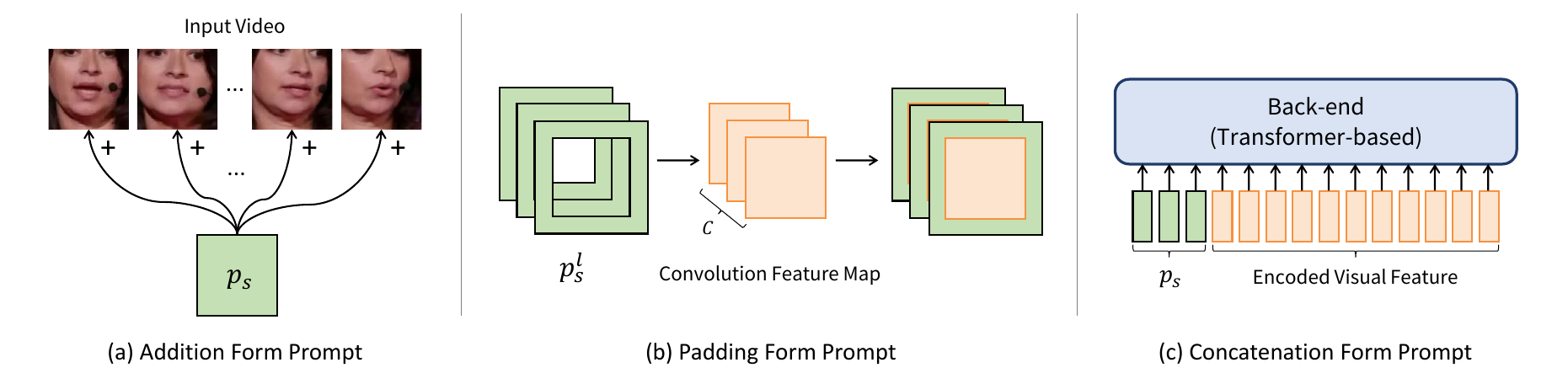}}
	\caption{Detailed illustrations of different types of prompt methods. (a) Addition form prompt is for being added to input video frames. (b) Padding form prompt is for being replaced the original padding region in the CNN. (c) Concatenation form prompt is for being concatenated to the input of Transformer-based module in the temporal dimension. Only prompts (\ie, green in the figure) are tuned during adaptation while maintaining the weight parameters of the pre-trained model.
	}
	\label{fig:2}
\end{figure*}
%#############################################
 
\section{Prompt Tuning for Speaker-adaptive Visual Speech Recognition}
As shown in Fig. \ref{fig:1}(a), VSR models usually consist of a visual front-end $\mathcal{F}$ which has CNN-based architecture, a back-end $\mathcal{B}$ which has RNN- or Transformer-based architecture, and predictor $\mathcal{P}$ that predicts the transcription. We denote all the learnable weight parameters of the VSR model including the visual front-end, the back-end, and the predictor as $\theta$. With a large-scale training dataset $\mathcal{T}=\{(\mathbf{X}^t, \mathbf{Y}^t)\}=\{(x^t_i, y^t_i)\}_{i=1}^{N_t}$ composed of $N_t$ pairs of input video $x^t_i$ and ground-truth text $y^t_i$, we can develop a VSR model having optimized parameter $\theta^*$, as follows,
\begin{align}
    \label{eq:1}
    \theta^* = \argmin_\theta\mathcal{L}(\mathbf{Y}^t, \hat{\mathbf{Y}}),
\end{align}
\begin{align}
    \label{eq:2}
    \hat{\mathbf{Y}} = (\mathcal{P}\circ \mathcal{B}\circ \mathcal{F})_\theta(\mathbf{X}^t),
\end{align}
where $\mathcal{L}$ is the objective function of VSR such as Connectionist Temporal Classification (CTC) \cite{graves2006ctc} and Cross Entropy, and $\circ$ denotes function composition.

Our objective is to maximize the performance of the pre-trained VSR model, $\theta^*$, on unseen speakers that did not appear in the training dataset $\mathcal{T}$. To this end, we utilize a small number of adaptation data $\mathcal{A}_s=\{(\mathbf{X}^{a_s}, \mathbf{Y}^{a_s})\}=\{(x^{a_s}_i, y^{a_s}_i)\}_{i=1}^{N_{a_s}}$ to adapt the pre-trained VSR model on the target speaker $s$. Here, we assume the amount of adaptation data is much smaller (\ie, 1$\sim$5 minutes) than the data used during training, $N_{a_s} \ll N_t$, which is the feasible data amount to obtain for each speaker in the real-world scenario. Specifically, we introduce prompts, $p$, and only optimize them to adapt the pre-trained model on unseen speakers instead of updating the learned weight parameters of the model. The proposed prompts have three different forms, addition, padding, and concatenation, which are illustrated in Fig. \ref{fig:1}(b). In the following subsections, we will describe details of how each prompt can be incorporated for developing speaker-adaptive VSR systems.

\subsection{Add Prompt: Addition to the input of CNN}
By adding some perturbations to the input of CNN, we can reprogram a pre-trained model to perform different tasks from those originally trained for. Motivated by these prior works \cite{elsayed2018AR1,chen2021AR2,dinh2022AR3,zheng2021AR4,neekhara2022AR5,yu2022defending}, we try to adapt a pre-trained VSR model to perform well on unseen speakers by adding a prompt to the input video. Therefore, only the prompt to be added to the input video is optimized by using a small amount of adaptation data of the target speaker. To this end, the prompt has the same size as the input video frame and it can be written as $p\in \mathbb{R}^{H\times W\times C}$, where the input video with $T$ frames can be written as $x_i^{a_s}\in \mathbb{R}^{T\times H\times W\times C}$. The prompt $p$ is added to the entire $T$ frames consistently and optimized to transform the input video of the target speaker to be operated well on the pre-trained VSR model. The optimization of the addition formula prompt for target speaker $s$ can be written as follows,
\begin{align}
    \label{eq:3}
    p_s^* = \argmin_{p_s}\mathcal{L}(\mathbf{Y}^{a_s}, \hat{\mathbf{Y}}),
\end{align}
\begin{align}
    \label{eq:4}
    \hat{\mathbf{Y}} = (\mathcal{P}\circ \mathcal{B}\circ \mathcal{F})_{\theta^*}(\mathbf{X}^{a_s} + p_s),
\end{align}
where the input of the pre-trained VSR model with parameter $\theta^*$ is now the video added with the prompt $p_s$ and the optimization is performed for the prompt while the model parameter $\theta^*$ is kept frozen. Please note that different from the adversarial examples \cite{goodfellow2014explaining}, the addition form prompt has no constraints with its value so that it can have a value out of range of the image similar to \cite{bahng2022visual}. The addition form prompt is illustrated in Fig. \ref{fig:2}(a). Since the learned weight parameters are maintained, we can naturally have the regularization effects \cite{li2006l2regularized,yu2013kldregularized} and it has a low risk for the model to be overfitted.

%------------------------------------ Table 1
%#############################################
\begin{table*}[t]
    \renewcommand{\arraystretch}{1.2}
	\centering
	\caption{Data information of 20 test speakers in LRW-ID.}
	\resizebox{0.97\linewidth}{!}{
	\begin{tabular}{cccccccccccc}
	\Xhline{3\arrayrulewidth}
	\textbf{Split} & \makecell{\textbf{Data} \\ \textbf{Information}} & \makecell{\textbf{S1} \\ (\#4243)} & \makecell{\textbf{S2} \\(\#5125)} & \makecell{\textbf{S3} \\(\#6003)} & \makecell{\textbf{S4}\\ (\#7184)} & \makecell{\textbf{S5} \\(\#9335)} & \makecell{\textbf{S6}\\ (\#9368)} & \makecell{\textbf{S7} \\(\#9438)} & \makecell{\textbf{S8} \\(\#9653)} & \makecell{\textbf{S9} \\(\#10209)} & \makecell{\textbf{S10}\\ (\#10293)} \\ \hline
    \multirow{3}{*}{Adapt.} 
        & Num. video & 565 & 743 & 1190 & 3271 & 2058 & 950 & 7239 & 622 & 745 & 738 \\ 
        & Data length (Min.) & 10.92 & 14.36 & 23.01 & 63.24 & 39.79 & 18.37 & 139.95 & 12.03 & 14.40 & 14.27 \\ 
        & Num. class & 252 & 329 & 425 & 473 & 418 & 346 & 493 & 282 & 316 & 290 \\ \hline
    \multirow{3}{*}{Test} 
        & Num. video & 565 & 743 & 1191 & 3271 & 2058 & 950 & 7239 & 623 & 745 & 739 \\
        & Data length (Min.) & 10.92 & 14.36 & 23.03 & 63.24 & 39.79 & 18.37 & 139.95 & 12.04 & 14.40 & 14.29 \\ 
        & Num. class & 240 & 313 & 416 & 476 & 412 & 346 & 495 & 290 & 330 & 294 \\ \hline
    \multirow{3}{*}{Total} 
        & Num. video & 1130 & 1486 & 2381 & 6542 & 4116 & 1900 & 14478 & 1245 & 1490 & 1477 \\ 
        & Data length (Min.) & 21.85 & 28.73 & 46.03 & 126.48 & 79.58 & 36.73 & 279.91 & 24.07 & 28.81 & 28.56 \\
        & Num. class & 316 & 402 & 478 & 494 & 453 & 421 & 497 & 365 & 411 & 358 \\ \hline
     \makecell{Adapt. \\ \& Test} & \makecell{Class overlap ratio \\(\%)} & 73.3 & 76.7 & 87.3 & 95.6 & 91.5 & 78.3 & 99.2 & 71.4 & 71.2 & 76.9 \\
    \Xhline{3\arrayrulewidth}
    \multicolumn{11}{c}{} \\ [-2.0ex]
    \Xhline{3\arrayrulewidth}
	\textbf{Split} & \makecell{\textbf{Data} \\ \textbf{Information}} & \makecell{\textbf{S11} \\ (\#10587)} & \makecell{\textbf{S12} \\(\#11041)} & \makecell{\textbf{S13} \\(\#11777)} & \makecell{\textbf{S14}\\ (\#11875)} & \makecell{\textbf{S15} \\(\#11910)} & \makecell{\textbf{S16}\\ (\#13287)} & \makecell{\textbf{S17} \\(\#13786)} & \makecell{\textbf{S18} \\(\#15545)} & \makecell{\textbf{S19} \\(\#15769)} & \makecell{\textbf{S20}\\ (\#17378)} \\ \hline
    \multirow{3}{*}{Adapt.} 
        & Num. Video & 553 & 2740 & 871 & 1400 & 476 & 606 & 827 & 2063 & 468 & 1793 \\ 
        & Data length (Min.) & 10.69 & 52.97 & 16.84 & 27.07 & 9.20 & 11.72 & 15.99 & 39.88 & 9.05 & 34.66 \\
        & Num. Class & 258 & 455 & 303 & 195 & 236 & 264 & 298 & 419 & 237 & 441 \\ \hline
    \multirow{3}{*}{Test} 
        & Num. Video & 553 & 2740 & 872 & 1400 & 476 & 607 & 827 & 2063 & 468 & 1793 \\
        & Data length (Min.) & 10.69 & 52.97 & 16.86 & 27.07 & 9.20 & 11.74 & 15.99 & 39.88 & 9.05 & 34.66 \\
        & Num. Class & 268 & 447 & 311 & 191 & 239 & 263 & 303 & 426 & 231 & 424 \\ \hline
    \multirow{3}{*}{Total} 
        & Num. Video & 1106 & 5480 & 1743 & 2800 & 950 & 1213 & 1654 & 4126 & 936 & 3586 \\ 
        & Data length (Min.) & 21.38 & 105.95 & 33.70 & 54.13 & 18.37 & 23.45 & 31.98 & 79.77 & 18.10 & 69.33 \\
        & Num. Class & 350 & 475 & 365 & 235 & 304 & 346 & 370 & 456 & 313 & 477 \\ \hline
     \makecell{Adapt. \\ \& Test} & \makecell{Class overlap ratio \\(\%)} & 65.7 & 95.5 & 80.1 & 79.1 & 71.5 & 68.8 & 76.2 & 91.3 & 67.1 & 91.5 \\
    \Xhline{3\arrayrulewidth}
	\end{tabular} 	\label{table:1}
	}
\end{table*}
%#############################################

\subsection{Replace Prompt: Padding of CNN}
Since the addition form prompt is applied at the input level, it can have insufficient representation power to transform the entire pre-trained model, especially if the model has large architecture. This problem is also shown in prompting in NLP, and previous work \cite{liu2021prompttuning} inserts the prompts into all layers of Transformer to enhance the representation power of the prompt. Taking inspiration from the prompting in NLP \cite{liu2021prompttuning}, we try to make the prompt can affect the intermediate layers of CNN. To this end, we propose to utilize the padding region of a convolution layer. Usually, padding is utilized in CNN to control the size of the output feature map, and to fill the padding region, zero, constant, and reflect values are employed. As the padding is also convolved with the learned convolution kernel, we replace these padding regions with the prompts to affect the encoded visual feature, and adapt the pre-trained VSR model to the unseen target speaker. Specifically, the prompts have the same size as the pre-defined padding size of each convolution layer. For example, the prompt for target speaker $s$ at $l$-th convolution layer can be written as $p_s^l \in \mathbb{R}^{S^l\times C^l}$, where $S^l=H^l(L^l+R^l)+W^l(U^l+B^l)+(L^l+R^l)(U^l+B^l)$ is the size of padding region, $L^l$, $R^l$, $B^l$, $U^l$ represent the padding size of left, right, bottom, and top, $H^l$, $W^l$, and $C^l$ are the height, width, and channel size of feature map of $l$-th layer. Then, the prompt tuning for the padding form can be written as follows,
\begin{align}
    \label{eq:5}
    \mathbf{P}_s^* = \argmin_{\mathbf{P}_s}\mathcal{L}(\mathbf{Y}^{a_s}, \hat{\mathbf{Y}}),
\end{align}
\begin{align}
    \label{eq:6}
    \hat{\mathbf{Y}} = (\mathcal{P}\circ \mathcal{B}\circ \mathcal{F})_{\theta^*}(\mathbf{X}^{a_s}, \mathcal{R}(\mathbf{P}_s)),
\end{align}
where $\mathbf{P}_s=\{p_s^l\}_1^{N_l}$ is the set of prompts for $N_l$ convolution layers and $\mathcal{R}(\cdot)$ is the operation of replacement the padding of visual front-end $\mathcal{F}$ with the prompts. Applying the padding form prompt is illustrated in Fig. \ref{fig:2}(b). Even if the padding is applied to the outer area of the feature map, it can affect the inner area of the feature map and even the entire feature map when the CNN layers are deep enough, with their enlarged receptive fields.

\subsection{Concatenate Prompt: Concatenation to Transformer input}
Recently, Transformer-based networks \cite{vaswani2017attention,gulati2020conformer} are proven to be more effective to model temporal data than RNN-based networks \cite{hochreiter1997lstm,chung2014gru,fragkiadaki2015recurrent}. Therefore, recently developed VSR models and state-of-the-art models \cite{ma2021end,afouras2018deep,hong2022visual,shi2022avhubert} are also utilizing Transformer-based back-end modules, and this makes it possible to utilize the prompting methods developed in NLP \cite{liu2021prompting,lester2021prompttuning} which are mainly for the Transformer-based architecture. Specifically, we concatenate the prompt to the input of back-end $\mathcal{B}$, the feature encoded from visual front-end $\mathcal{F}$, in the temporal dimension. The prompt has the same dimension as that of the encoded feature and can be denoted as $p_s\in\mathbb{R}^{N_p\times D}$, where $N_p$ is the length of prompts and $D$ is the dimension of the encoded feature. Then, through the self-attention layers in the back-end, the prompts can affect the features encoded by the back-end. The optimization of concatenation form prompts can be written as follows,
\begin{align}
    \label{eq:7}
    p_s^* = \argmin_{p_s}\mathcal{L}(\mathbf{Y}^{a_s}, \hat{\mathbf{Y}}),
\end{align}
\begin{align}
    \label{eq:8}
    \hat{\mathbf{Y}} = (\mathcal{P}\circ \mathcal{B})_{\theta^*}(\mathcal{F}(\mathbf{X}^{a_s})_{\theta^*} \oplus p_s),
\end{align}
where $\oplus$ represents concatenation in the temporal dimension. The concatenation form prompt is illustrated in Fig. \ref{fig:2}(c).

The three types of prompts, addition form, padding form, and concatenation form, can be jointly utilized for a pre-trained VSR model to adapt the model to an unseen target speaker. We evaluate the effectiveness of each prompt method and the combination of them through extensive experiments in the following sections.

\section{Experimental Setup}
For the experiments, we utilize both word- and sentence-level audio-visual datasets, LRW \cite{chung2016lrw} and GRID \cite{cooke2006grid}. We evaluate the proposed method by adapting a pre-trained model on unseen speakers by using different amounts of adaptation data. The following subsections describe the details of data settings and implementation details for each dataset.
\subsection{Dataset}
\subsubsection{GRID}
GRID \cite{cooke2006grid} is a sentence-level audio-visual corpus dataset following a fixed grammar. It consists of videos uttered by 33 speakers. Each speaker contains about 1,000 videos and each video is 3 seconds long. We follow the unseen speaker split of \cite{assael2016lipnet} so that speakers 1, 2, 20, and 22 are used for testing while the remainder is utilized for training. The data of the test speakers are divided into half to build adaptation sets and test sets. Since each video is 3 seconds long, every 20 videos from the adaptation set compose 1 minute of adaptation data. For the pre-processing and data augmentation, the lip ROI is cropped and resized into a size of 64$\times$128 similar to \cite{assael2016lipnet}, and random horizontal flipping, time masking, and random spatial region erasing are employed similar to \cite{kim2022distinguishing}. For the performance metric, Word Error Rate (WER, \%) which measures the prediction error compared to ground truth is utilized for the dataset. Therefore, the lower WER values indicate better VSR performances.

\subsubsection{LRW-ID}
LRW \cite{chung2016lrw} is a word-level audio-visual corpus dataset captured in television programs, thus datasets have large pose and illumination variations that are close to the real-world setting. It consists of 500 word classes and each word class contains 1,000 training videos. Since the dataset doesn't contain the speaker information, we annotate the speaker by using face recognition \cite{barr2012face,he2005face} technology with a pre-trained model \cite{deng2019arcface}. The annotated speakers are 17,580 which is very large compared to the GRID dataset. To build the speaker-adaptation setting, we set 20 speakers for testing while the remainder is utilized for training, and named the modified split as LRW-ID. The data information of 20 test speakers of LRW-ID is shown in Table \ref{table:1}. S\# represents the speaker index and the number in the blanket represents the speaker ID. Compared to GRID dataset, each speaker has different amounts of adaptation data, and the adaptation set may not cover the entire word classes in the test set, which is closer to a real-world scenario. Since each video is 1.16 seconds long, 52, 155, and 259 videos compose 1, 3, and 5 minutes of adaptation data. The lip ROI is cropped and resized into 112$\times$112, and the cropped frames are converted into grayscale. The same data augmentation is applied as GRID dataset. For the performance metric, word accuracy (ACC, \%) is employed so the large value of ACC refers to better VSR performance.

%------------------------------------ Table 2
%#############################################
\begin{table}[t]
\caption{Network architecture for GRID dataset.}
    \centering
	\renewcommand{\arraystretch}{1.5}
	\renewcommand{\tabcolsep}{1.0mm}
    \resizebox{0.999\linewidth}{!}{
    \begin{tabular}{cccc}
    \Xhline{3\arrayrulewidth}
    \multicolumn{4}{c}{\textbf{Input size}: 75 $\times$ 64 $\times$ 128 $\times$ 3 (T $\times$ H $\times$ W $\times$ C)} \\\Xhline{3\arrayrulewidth}
    \textbf{Layer}  & \textbf{Filter size / number / stride}  & \textbf{Padding Prompt}& \textbf{Output dimensions} \\ \hline
    Conv 3D & 3 $\times$ 5 $\times$ 5 / 32 / [1, 2, 2] & [2, 2] $\times$ 3 & 75$\times$32$\times$64$\times$32  \\ \hline
    Maxpool & 2 $\times$ 2 / - / [2, 2] & - & 75$\times$16$\times$32$\times$32\\ \hline
    Conv 3D & 3 $\times$ 5 $\times$ 5 / 64 / [1, 1, 1] & [2, 2] $\times$ 32 & 75$\times$8$\times$16$\times$64 \\ \hline
    Maxpool & 2 $\times$ 2 / - / [2, 2] & - & 75$\times$4$\times$8$\times$64\\ \hline
    Conv 3D & 3 $\times$ 3 $\times$ 3 / 96 / [1, 1, 1] & [1, 1] $\times$ 64 & 75$\times$4$\times$8$\times$96 \\ \hline
    Maxpool & 2 $\times$ 2 / - / [2, 2] & - & 75$\times$2$\times$4$\times$96\\ \hline
    Conv 2D & 3 $\times$ 3 / 32 / [2, 2] & [1, 1] $\times$ 96 & 75$\times$2$\times$4$\times$32 \\ \hline
    Conv 2D & 3 $\times$ 3 / 64 / [2, 2] & [1, 1] $\times$ 32 & 75$\times$1$\times$2$\times$64 \\ \hline
    Flatten & - & - & 75$\times$128\\ \hline
    Linear & 128 $\times$ 256 & - & 75$\times$256\\ \hline
    Transformer & 256 / 4 layers & - & 75$\times$256\\ \hline
    Linear & 256 $\times$ Num\_class & - & 75$\times$Num\_class\\ \Xhline{3\arrayrulewidth}
    \end{tabular}}
	\label{table:2}
\end{table}
%#############################################

%------------------------------------ Table 3
%#############################################
\begin{table}[t!]
\caption{Network architecture for LRW-ID dataset.}
    \centering
	\renewcommand{\arraystretch}{1.7}
	\renewcommand{\tabcolsep}{1.0mm}
    \resizebox{0.999\linewidth}{!}{
    \begin{tabular}{cccc}
    \Xhline{3\arrayrulewidth}
    \multicolumn{4}{c}{\textbf{Input size}: 29 $\times$ 112 $\times$ 112 $\times$ 1 (T $\times$ H $\times$ W $\times$ C)} \\\Xhline{3\arrayrulewidth}
    \textbf{Layer}  & \textbf{Filter size / number / stride}  & \textbf{Padding Prompt}& \textbf{Output dimensions} \\ \hline
    Conv 3D & 5 $\times$ 7 $\times$ 7 / 64 / [1, 2, 2] & [3, 3] $\times$ 1 & 29$\times$64$\times$64$\times$64 \\ \hline
    Max Pool & 3 $\times$ 3 / - / [2, 2] & - & 29$\times$32$\times$32$\times$64\\ \hline
    ResBlock & \makecell{3 $\times$ 3 / 64 / [1, 1] \\ 3 $\times$ 3 / 64 / [1, 1]} & \makecell{$\left[1, 1\right]$\, $\times$ 64 \\ $\left[1, 1\right]$\, $\times$ 64} & 29$\times$32$\times$32$\times$64 \\ \hline
    ResBlock & \makecell{3 $\times$ 3 / 64 / [1, 1] \\ 3 $\times$ 3 / 64 / [1, 1]} & \makecell{$\left[1, 1\right]$\, $\times$ 64 \\ $\left[1, 1\right]$\, $\times$ 64} & 29$\times$32$\times$32$\times$64 \\ \hline
    
    ResBlock & \makecell{3 $\times$ 3 / 128 / [2, 2] \\ 3 $\times$ 3 / 128 / [1, 1]} & \makecell{$\left[1, 1\right]$\, $\times$ 64 \\ $\left[1, 1\right]$\, $\times$ 128} & 29$\times$16$\times$16$\times$128 \\ \hline
    ResBlock & \makecell{3 $\times$ 3 / 128 / [1, 1] \\ 3 $\times$ 3 / 128 / [1, 1]} & \makecell{$\left[1, 1\right]$\, $\times$ 128 \\ $\left[1, 1\right]$\, $\times$ 128} & 29$\times$16$\times$16$\times$128 \\ \hline
    
    ResBlock & \makecell{3 $\times$ 3 / 256 / [2, 2] \\ 3 $\times$ 3 / 256 / [1, 1]} &  \makecell{$\left[1, 1\right]$\, $\times$ 128 \\ $\left[1, 1\right]$\, $\times$ 256} & 29$\times$8$\times$8$\times$256 \\ \hline
    ResBlock & \makecell{3 $\times$ 3 / 256 / [1, 1] \\ 3 $\times$ 3 / 256 / [1, 1]} &  \makecell{$\left[1, 1\right]$\, $\times$ 256 \\ $\left[1, 1\right]$\, $\times$ 256} & 29$\times$8$\times$8$\times$256 \\ \hline
    
    ResBlock & \makecell{3 $\times$ 3 / 512 / [2, 2] \\ 3 $\times$ 3 / 512 / [1, 1]} &  \makecell{$\left[1, 1\right]$\, $\times$ 256 \\ $\left[1, 1\right]$\, $\times$ 512} & 29$\times$4$\times$4$\times$512 \\ \hline
    ResBlock & \makecell{3 $\times$ 3 / 512 / [1, 1] \\ 3 $\times$ 3 / 512 / [1, 1]} &  \makecell{$\left[1, 1\right]$\, $\times$ 512 \\ $\left[1, 1\right]$\, $\times$ 512} & 29$\times$4$\times$4$\times$512 \\ \hline
    Flatten & - & - & 29$\times$8192 \\ \hline
    Linear & 8192 $\times$ 512 & - & 29$\times$512 \\ \hline
    Transformer & 512 / 6 layers & - & 29$\times$512 \\ \hline
    Temp. Avg. & - & - & 512 \\ \hline
    Linear & 512 $\times$ Num\_class & - & Num\_class \\
    \Xhline{3\arrayrulewidth}
    \end{tabular}}
	\label{table:3}
\end{table}
%#############################################

\subsection{Implementation details}
The basic architecture of VSR models is similar as illustrated in Fig. \ref{fig:1}(a). For the GRID dataset, we modify the architecture of \cite{assael2016lipnet}. The visual front-end is composed of three 3D convolutions and two 2D convolutions, the back-end is composed of a 4-layered Transformer \cite{vaswani2017attention} with a hidden dimension size of 256, and the predictor is composed of a linear layer (Table \ref{table:2}). For training, we use word-level CTC \cite{graves2006ctc} loss function. The size of the addition prompt for GRID dataset is 64$\times$124$\times$3 which is the same as the input frame, and it is added to all input frames consistently. To insert the padding prompt, all paddings in 5 convolution layers are changed from zero paddings to the padding prompts. The detailed size of convolution layers and prompts are shown in Table \ref{table:2}. The padding prompt size is represented as [left and right padding size, top and bottom padding size] $\times$ channel size. The concatenation prompt is concatenated with the encoded visual feature before passing the back-end module in the time dimension. We use 5 for the length of the concatenation prompt (\ie, $N_p=5$).

For the LRW-ID dataset, we use the ResNet-18 architecture \cite{he2016resnet} whose first convolution layer is changed with 3D convolution following \cite{petridis2018end,kim2021cromm}. For the back-end module, we employ a 6-layered Transformer with a hidden size of 512, and a linear layer is utilized for the predictor (Table \ref{table:3}). For the loss function, Cross Entropy loss is applied for 500 word classes. For the LRW-ID, the addition prompt has the size of 112$\times$112$\times$1, the padding prompt is inserted for all convolution layers (\ie, 17 layers) in the visual front-end as described in Table \ref{table:3}, and the concatenation prompt is inserted before the back-end, with 5 lengths.

%------------------------------------ Table 4
%#############################################
\begin{table}[t]
	\renewcommand{\arraystretch}{1.2}
	\renewcommand{\tabcolsep}{4mm}
    \centering
    \caption{Number of parameters of each prompt for one speaker}
    \resizebox{0.999\linewidth}{!}{
	\begin{tabular}{ccc}
	\Xhline{3\arrayrulewidth}
    \textbf{Method} & \textbf{GRID} & \textbf{LRW-ID} \\ \hline
    Add & 24.58K (0.687\%) & 12.54K (0.036\%) \\
    Pad & 15.54K (0.434\%) & 153.19K (0.443\%) \\
    Cat & 1.28K (0.036\%) & 2.56K (0.007\%) \\
    Add + Pad & 40.11K (1.121\%) & 165.73K (0.480\%) \\
    Add + Cat & 25.86K (0.722\%) & 15.10K (0.044\%) \\
    Pad + Cat & 16.82K (0.470\%) & 155.75K (0.451\%) \\
    Add + Pad + Cat & 41.39K (1.156\%) & 168.29K (0.487\%) \\
	\Xhline{3\arrayrulewidth}
	\end{tabular}  \label{table:4}
	}
\end{table}
%#############################################

For evaluating the proposed prompt tuning method in speaker adaptation, we pre-train the VSR models on the training set whose subjects are not overlapped with the testing set. For pre-training on GRID, a batch size of 112 and a maximum learning rate of 0.008 with 5,000 warmup \cite{vaswani2017attention} steps are used. The training data consists of about 29,000 videos. For LRW-ID, a batch size of 400 and a maximum learning rate of 0.004 with 10,000 warmup steps are used, and the training data consists of 480,378 videos. With the pre-trained model, $\theta^*$, we only optimize the prompts to adapt the model to unseen speakers while the model parameter is kept frozen. During adaptation, we use a learning rate of 0.01 for the addition and padding prompts and a learning rate of 0.1 for the concatenation prompt, with a batch size of 112 and 55 for GRID and LRW-ID, respectively. For the optimization, we use AdamW \cite{kingma2014adam,loshchilov2017adamw} and TITAN RTX GPUs for both pre-train and adaptation. The number of parameters for each prompt method for adapting one target speaker is shown in Table \ref{table:4}. The percentage refers to the relative number of parameters compared to the full model parameters. It shows that the prompt has a very small number of parameters compared to the full model.

%------------------------------------ Table 5
%#############################################
\begin{table}[t]
    \renewcommand{\arraystretch}{1.2}
    \renewcommand{\tabcolsep}{3.0mm}
	\centering
	\caption{Adaptation results (WER) on GRID using 1, 3, and 5 minutes of adaptation data}
	\resizebox{0.999\linewidth}{!}{
	\begin{tabular}{ccccccc}
	\Xhline{3\arrayrulewidth}
	\textbf{Method} & \makecell{\textbf{Adapt.}\\\textbf{min}}
	& \textbf{S1} & \textbf{S2} & \textbf{S20} & \textbf{S22} & \textbf{Mean}\\ \hline
    Baseline \cite{assael2016lipnet} & - & 16.40 & 9.42 & 11.23 & 11.57 & 12.04 \\ \hline
    \multirow{3}{*}{Add} & 1 & 8.08 & 3.74 & 5.83 & 4.33 & 5.48 \\
                         & 3 & 6.36 & 2.54 & 5.23 & 3.17 & 4.31\\
                         & 5 & 5.99 & 2.47 & 5.27 & 3.03 & 4.17\\ \hline
    \multirow{3}{*}{Pad} & 1 & 10.61 & 3.84 & 6.83 & 4.53 & 6.41 \\
                         & 3 & 8.72 & 2.57 & 6.03 & 3.87 & 5.28 \\
                         & 5 & 7.81 & 2.61 & 5.87 & 3.77 & 4.99 \\ \hline
    \multirow{3}{*}{Cat} & 1 & 11.79 & 4.14 & 7.63 & 5.37 & 7.21 \\
                         & 3 & 9.43 & 3.94 & 6.47 & 3.90 & 5.92 \\
                         & 5 & 8.25 & 3.37 & 6.27 & 3.50 & 5.32 \\ \hline
    \multirow{3}{*}{\makecell{Add\\Pad}} 
                         & 1 & 8.65 & 4.01 & 6.43 & 4.93 & 5.97 \\
                         & 3 & 7.10 & 2.57 & 5.60 & 3.37 & 4.65 \\
                         & 5 & 5.79 & 2.37 & \textbf{5.13} & 2.97 & 4.04 \\ \hline
    \multirow{3}{*}{\makecell{Add\\Cat}} 
                         & 1 & 7.81 & 3.24 & 5.80 & 4.20 & 5.25 \\
                         & 3 & 6.16 & 2.34 & 5.63 & 3.27 & 4.33 \\
                         & 5 & 5.12 & 2.31 & 5.63 & 2.97 & 4.00 \\ \hline   
    \multirow{3}{*}{\makecell{Pad\\Cat}} 
                         & 1 & 11.45 & 3.77 & 6.97 & 4.27 & 6.60 \\
                         & 3 & 8.52 & 2.74 & 6.20 & 3.80 & 5.31 \\
                         & 5 & 6.30 & 2.47 & 5.53 & 3.27 & 4.37 \\ \hline    
    \multirow{3}{*}{\makecell{Add\\Pad\\Cat}} 
                         & 1 & 7.91 & 3.81 & 6.07 & 4.43 & 5.53 \\
                         & 3 & 6.43 & \textbf{2.14} & 5.63 & 3.07 & 4.31 \\
                         & 5 & \textbf{5.08} & 2.24 & \textbf{5.13} & \textbf{2.80} & \textbf{3.80} \\
    \Xhline{3\arrayrulewidth}
    
	\end{tabular} 	\label{table:5}
	}
\end{table}
%#############################################

\subsection{Baselines for comparisons}
In order to validate the effectiveness of the proposed prompt tuning methods in speaker-adaptive VSR, we set comparison methods including previous speaker-adaptive methods and finetuning methods. 

\textit{Baseline} is a pre-trained VSR model without performing the adaptation to the unseen speakers so it shows the lower bound performance.
\textit{Speaker-invariant}\cite{meng2018sitASR} and \textit{Speaker code}\cite{abdel2013codeicassp} are the speaker-invariant and -adaptive methods developed for ASR models. We directly apply these methods to the VSR model to compare the effectiveness of the proposed method with previous speaker-invariant and -adaptive methods. Specifically, for the speaker-invariant model, we additionally attach a speaker identity classifier after the visual front-end. During training, the speaker identity classifier is guided to classify the subject identity from the encoded visual feature while the visual front-end module is guided to deceive the speaker identity classifier. Therefore, with the adversarial training \cite{ganin2015grl}, the visual front-end is eventually become a speaker-invariant model by not encoding the speaker identity information into the visual feature. For the speaker-adaptive model, \textit{Speaker code}, we additionally train the Adaptation Network and speaker code of \cite{abdel2013codeicassp} on training dataset $\mathcal{T}$ by attaching them to the pre-trained VSR model. After training, adaptation is performed by only training the speaker code on the adaptation dataset of target speaker $s$, $\mathcal{A}_s$. We use 128, 64, and 32 dimensions of speaker code for each MLP layer in the Adaptation Network for GRID dataset, and 256, 128, and 64 dimensions for LRW-ID dataset.

Moreover, to compare with the finetuning methods \cite{liao2013trainpart1}, we set three types of finetuning methods by differing the trainable parts of the pre-trained model. Firstly, \textit{FineTune-C} is the method that only the last linear layer (\ie, predictor) is tuned on the adaptation data, so it has the smallest number of parameters compared to other finetuning methods. \textit{FineTune-B} is the method for finetuning both the back-end and the predictor. Finally, \textit{FineTune-F} is the method that the whole pre-trained VSR model is finetuned on the adaptation dataset, hence it requires the largest parameters. For finetuning the pre-trained model, a learning rate of 1e-5 is utilized.

\section{Experimental Results}
We evaluate the effectiveness of the proposed prompt tuning in speaker-adaptive VSR by using different amounts of adaptation data. Firstly, we explore the effectiveness of the proposed method in a situation where only a small number of adaptation data is available. To this end, we use 1, 3, and 5 minutes of adaptation data for each unseen speaker to adapt the pre-trained model to the target speaker. Then, we examine the speaker adaptation performances by using different ratios of adaptation data, to validate when the prompt tuning has benefits compared to finetuning methods.

\subsection{Speaker adaptation results with small data}
In this experiment, we evaluate the effectiveness of the different types of prompts, the addition, the padding, the concatenation, and their combinations by using a small number of adaptation data with 1, 3, and 5 minutes in length. Therefore, a total of 7 methods are evaluated including, addition prompt only (Add), padding prompt only (Pad), concatenation prompt only (Cat), and their combinations (Add Pad), (Add Cat), (Pad Cat), and (Add Pad Cat). The speaker adaptation results by using the proposed prompt tuning for 4 unseen test speakers (\ie, S1, S2, S20, and S22) on GRID are shown in Table \ref{table:5}. Compared to the performance of Baseline which is the results obtained by using the pre-trained VSR model on unseen target speakers directly, all prompt methods largely improve the performance on the target unseen speaker. The results indicate that the VSR model is inherently sensitive to personal lip appearances and a pre-trained VSR model can show degraded performance when they are applied to unseen speakers directly. In contrast, by applying the proposed speaker adaptation method, we can improve the VSR performances with a small number of adaptation data, so that it achieves the almost similar performance obtained in seen speaker settings \cite{assael2016lipnet,chung2017lrs2}. Please note that just using 1 minute of adaptation data improves 56\% relative mean performance with the addition and concatenation prompts (Add Cat), compared to Baseline.

%------------------------------------ Table 6
%#############################################
\begin{table*}[t]
    \renewcommand{\arraystretch}{1.2}
    \renewcommand{\tabcolsep}{3.0mm}
	\centering
	\caption{Mean adaptation results (ACC) of 20 speakers on LRW-ID using 1, 3, and 5 minutes of adaptation data}
	\resizebox{0.7\linewidth}{!}{
	\begin{tabular}{cccccccc|c}
	\Xhline{3\arrayrulewidth}
    \textbf{Adapt. min} & \textbf{Add} & \textbf{Pad} & \textbf{Cat} & \textbf{A+P} & \textbf{A+C} & \textbf{P+C} & \textbf{A+P+C} & \textbf{Baseline} \\ \hline
    1 & 87.75 & 88.42 & 88.28 & 88.19 & 88.06 & \underline{88.53} & \textbf{88.55} & \multirow{3}{*}{87.54} \\
    3 & 87.89 & 89.32 & 88.92 & 88.88 & 88.69 & \textbf{89.45} & \underline{89.39} & \\
    5 & 88.03 & 89.62 & 89.33 & 89.14 & 89.02 & \textbf{89.99} & \underline{89.75} & \\
    \Xhline{3\arrayrulewidth}
	\end{tabular} 	\label{table:6}
	}
\end{table*}
%#############################################

By comparing the different prompt methods, the addition prompt achieves the best performance among the single prompt methods for the GRID dataset. It is related to the used network architecture of GRID shown in Table \ref{table:2}, which has a shallow visual front-end so the input-level prompt can affect the visual feature largely, while the padding prompt has fewer effects on the inner feature map due to the shallow layers and small receptive fields. Moreover, by additionally utilizing the concatenation prompt with the prompts for CNN (\ie, addition and padding prompts), the VSR performances on the target unseen speakers are improved overall. Analyzing the effect of each prompt, the addition and the padding prompts are mainly for improving the encoded visual feature so that the target speaker's personal lip appearances can be adaptively modeled, and the concatenation prompts improves the temporal encoding availability of the back-end module so that the personal lip movements can be accounted. We find that the combination of the addition and the padding prompts, (Add Pad), does not further improve the performance over the addition-only prompts (Add). A similar tendency can be seen in that the (Add Cat) and (Add Pad Cat) have similar performances.

The speaker adaptation results for 20 unseen test speakers on LRW-ID are shown in Table \ref{table:7}. Even if the pre-trained VSR model is trained with large speaker variations with 17,560 speakers, we can still improve the VSR performance by adapting the pre-trained model on the target unseen speaker. Especially, when the model has less competency for a test speaker, the effectiveness of adaptation is bigger. For example, the pre-trained VSR model shows 75.95\% ACC on speaker 11 (S11) which is lower compared to the mean ACC of 87.54\% over all speakers. In this case, by using the proposed prompt tuning (Pad Cat) with 3 minutes of adaptation data, we can improve the performance to 84.45\% ACC which is a large improvement of about 8.5\% ACC. On the other hand, we find that when the pre-trained model has enough competency, the effectiveness of the adaptation is small. This can be seen from the results of speaker 9 (S9), where the performance gain is smaller compared to the other speakers. As the purpose of the speaker adaptation is to transform a pre-trained model that cannot capture the lip appearances and movements of the target unseen speaker, the obtained tendency is natural and the effect of the adaptation can be small for the already well-captured lip appearances and movements. 

%------------------------------------ Table 7
%#############################################
\begin{table*}[h!]
    \renewcommand{\arraystretch}{1.2}
	\centering
	\caption{Adaptation results (ACC) on LRW-ID using 1, 3, and 5 minutes of adaptation data}
	\resizebox{0.97\linewidth}{!}{
	\begin{tabular}{cccccccccccc}
	\Xhline{3\arrayrulewidth}
	\textbf{Method} & \makecell{\textbf{Adapt.}\\\textbf{min}}
	& \makecell{\textbf{S1} \\ (\#4243)} & \makecell{\textbf{S2} \\(\#5125)} & \makecell{\textbf{S3} \\(\#6003)} & \makecell{\textbf{S4}\\ (\#7184)} & \makecell{\textbf{S5} \\(\#9335)} & \makecell{\textbf{S6}\\ (\#9368)} & \makecell{\textbf{S7} \\(\#9438)} & \makecell{\textbf{S8} \\(\#9653)} & \makecell{\textbf{S9} \\(\#10209)} & \makecell{\textbf{S10}\\ (\#10293)} \\ \hline
    Baseline & - & 75.75 & 84.39 & 84.80 & 90.40 & 82.70 & 84.74 & 91.95 & 82.18 & 90.60 & 81.46 \\ \hline
    \multirow{3}{*}{Add} & 1 & 76.81 & 84.39 & 84.80 & 90.40 & 82.70 & 84.74 & 91.95 & 82.18 & 90.60 &                      82.41 \\
                         & 3 & 77.52 & 84.93 & 84.80 & 90.65 & 82.70 & 84.74 & 91.95 & 82.18 & 90.60 & 83.09 \\
                         & 5 & 78.94 & 85.33 & 85.14 & 90.49 & 82.99 & 84.74 & 92.06 & 82.99 & 90.60 & 83.63 \\ \hline
    \multirow{3}{*}{Pad} & 1 & 79.47 & 85.60 & 86.06 & 90.40 & 85.33 & 84.74 & 92.04 & 82.83 & 90.60 &                      84.84 \\
                         & 3 & 82.30 & 86.03 & 86.48 & 90.83 & 85.52 & 86.21 & 93.31 & 84.59 & 90.60 & 84.57 \\
                         & 5 & 82.30 & 86.68 & \textbf{87.74} & 90.98 & 86.59 & 86.32 & 93.18 & 84.43 & 90.60 & 86.33 \\ \hline
    \multirow{3}{*}{Cat} & 1 & 78.94 & 84.39 & 84.80 & 90.49 & 85.47 & 84.74 & 91.95 & 82.34 & 90.60 &                      82.41 \\
                         & 3 & 82.30 & 84.93 & 86.06 & 90.58 & 85.33 & 85.05 & 92.58 & 83.31 & \textbf{90.87} & 84.44 \\
                         & 5 & 82.48 & 86.14 & 86.57 & 90.74 & 86.25 & 85.58 & 93.02 & \textbf{85.39} & \textbf{90.87} & 85.12 \\ \hline
    \multirow{3}{*}{\makecell{Add\\Pad}} 
                         & 1 & 79.82 & 84.79 & 84.80 & 90.40 & 84.79 & 84.74 & 91.95 & 82.83 & 90.60 & 83.36 \\
                         & 3 & 81.59 & 86.54 & 86.06 & 90.86 & 84.60 & 85.68 & 92.65 & 83.95 & 90.60 & 84.71 \\
                         & 5 & 82.12 & 86.95 & 87.07 & 90.89 & 85.47 & 86.63 & 92.67 & 83.95 & 90.60 & 86.20 \\ \hline
    \multirow{3}{*}{\makecell{Add\\Cat}} 
                         & 1 & 79.29 & 84.52 & 84.80 & 90.40 & 84.11 & 84.74 & 91.95 & 82.99 & 90.60 & 81.87 \\
                         & 3 & 81.95 & 85.33 & 86.40 & 90.40 & 84.16 & 85.37 & 92.68 & 82.99 & 90.60 & 83.90 \\
                         & 5 & 82.66 & 85.20 & 86.57 & 90.55 & 84.94 & 86.53 & 93.04 & 84.59 & 90.60 & 84.84 \\ \hline   
    \multirow{3}{*}{\makecell{Pad\\Cat}} 
                         & 1 & 80.53 & 84.79 & 85.39 & 90.40 & 86.05 & 84.74 & 92.07 & 83.15 & \textbf{90.87} & 84.98 \\
                         & 3 & 82.66 & 86.41 & 86.90 & 90.68 & 86.40 & 86.00 & 93.19 & 84.91 & \textbf{90.87} & 85.25 \\
                         & 5 & 84.07 & 86.95 & 87.49 & \textbf{91.17} & \textbf{87.12} & 86.84 & \textbf{93.56} & \textbf{85.39} & 90.74 & \textbf{87.28} \\ \hline    
    \multirow{3}{*}{\makecell{Add\\Pad\\Cat}} 
                         & 1 & 81.06 & 85.73 & 85.31 & 90.40 & 85.81 & 84.74 & 92.00 & 83.95 & \textbf{90.87} & 84.03 \\
                         & 3 & 82.66 & 86.14 & 87.41 & 90.52 & 86.40 & 86.21 & 92.83 & 85.71 & 90.60 & 85.25 \\
                         & 5 & \textbf{84.78} & \textbf{87.21} & 87.32 & 91.10 & \textbf{87.12} & \textbf{87.68} & 93.26 & \textbf{85.39} & 90.60 & 86.33 \\
    \Xhline{3\arrayrulewidth}
    \multicolumn{12}{c}{} \\ [-1.2ex]
    \Xhline{3\arrayrulewidth}
	\textbf{Method} & \makecell{\textbf{Adapt.}\\\textbf{min}}
	& \makecell{\textbf{S11} \\ (\#10587)} & \makecell{\textbf{S12} \\(\#11041)} & \makecell{\textbf{S13} \\(\#11777)} & \makecell{\textbf{S14}\\ (\#11875)} & \makecell{\textbf{S15} \\(\#11910)} & \makecell{\textbf{S16}\\ (\#13287)} & \makecell{\textbf{S17} \\(\#13786)} & \makecell{\textbf{S18} \\(\#15545)} & \makecell{\textbf{S19} \\(\#15769)} & \makecell{\textbf{S20}\\ (\#17378)}  \\ \hline
    Baseline & - & 75.95 & 88.03 & 88.30 & 89.71 & 75.21 & 75.78 & 82.59 & 89.77 & 90.60 & 89.01 \\ \hline
    \multirow{3}{*}{Add} & 1 & 77.76 & 88.29 & 88.88 & 90.21 & 75.21 & 76.94 & 82.59 & 89.82 & 90.60 &                      89.52 \\
                         & 3 & 78.12 & 88.07 & 89.11 & 92.00 & 75.21 & 76.77 & 82.59 & 89.87 & 90.60 & 89.52 \\
                         & 5 & 78.84 & 88.25 & 88.99 & 91.14 & 75.21 & 77.43 & 82.59 & 89.87 & 90.60 & 89.96 \\ \hline
    \multirow{3}{*}{Pad} & 1 & 81.56 & 88.03 & 89.11 & 92.00 & 76.89 & 80.56 & 83.92 & 89.77 & 90.60 &                      89.52 \\
                         & 3 & 85.35 & 89.75 & 90.48 & 92.50 & 78.57 & 79.74 & 84.89 & 89.77 & 90.81 & 90.18 \\
                         & 5 & 83.73 & \textbf{90.58} & 90.60 & 92.64 & 77.94 & 80.56 & 85.49 & 89.87 & 90.60 & 90.97 \\ \hline
    \multirow{3}{*}{Cat} & 1 & 79.02 & 88.50 & 89.68 & 91.64 & 78.15 & 78.58 & 82.71 & 90.11 & \textbf{91.67} &                      90.18 \\
                         & 3 & 82.64 & 89.05 & 90.14 & 92.86 & 79.41 & 79.57 & 83.07 & 90.26 & 91.24 & 90.69 \\
                         & 5 & 82.82 & 89.67 & 89.79 & 92.57 & 78.78 & 80.56 & 84.04 & \textbf{90.45} & 91.45 & 90.46 \\ \hline
    \multirow{3}{*}{\makecell{Add\\Pad}} 
                         & 1 & 81.01 & 88.03 & 89.11 & 91.50 & 76.05 & 78.58 & 84.16 & 89.77 & 90.60 & 89.79 \\
                         & 3 & 81.56 & 89.56 & 90.25 & 92.14 & 78.15 & 78.42 & 84.28 & 89.77 & 90.60 & 90.13 \\
                         & 5 & 81.56 & 89.34 & 90.37 & 92.43 & 78.78 & 78.58 & 84.16 & 90.02 & 90.60 & 90.63 \\ \hline
    \multirow{3}{*}{\makecell{Add\\Cat}} 
                         & 1 & 79.02 & 88.39 & 89.45 & 90.86 & 75.63 & 78.75 & 82.59 & 90.16 & 90.60 & 89.96 \\
                         & 3 & 80.11 & 89.02 & 90.02 & 92.43 & 78.15 & 78.91 & 82.95 & 90.31 & 91.24 & 90.69 \\
                         & 5 & 81.56 & 89.05 & 89.79 & 92.07 & 77.73 & 80.89 & 84.16 & 90.11 & 90.60 & 90.91 \\ \hline   
    \multirow{3}{*}{\makecell{Pad\\Cat}} 
                         & 1 & 81.01 & 88.03 & 89.56 & 92.43 & 77.10 & 80.40 & 84.16 & 89.77 & 90.81 & 90.07 \\
                         & 3 & \textbf{84.45} & 89.85 & \textbf{90.71} & 92.64 & 80.04 & 80.07 & 84.89 & 89.77 & 90.60 & 90.80 \\
                         & 5 & 83.36 & 90.51 & 90.37 & 93.07 & \textbf{80.46} & 82.04 & \textbf{85.85} & 90.40 & 90.60 & \textbf{91.19} \\ \hline    
    \multirow{3}{*}{\makecell{Add\\Pad\\Cat}} 
                         & 1 & 81.01 & 88.03 & 89.56 & 92.29 & 78.15 & 80.07 & 83.80 & 90.21 & 90.81 & 90.13 \\
                         & 3 & \textbf{84.45} & 90.04 & \textbf{90.71} & \textbf{93.36} & 78.99 & 80.23 & 83.92 & 89.82 & 90.60 & 90.69 \\
                         & 5 & 82.10 & 89.82 & 90.37 & 92.86 & 80.25 & \textbf{82.37} & 84.89 & 89.82 & 90.60 & 90.91 \\
    \Xhline{3\arrayrulewidth}
	\end{tabular} 	\label{table:7}
	}
\end{table*}
%#############################################

The mean word accuracies for 20 speakers (\ie, listed in Table \ref{table:7}) according to different types of prompts are presented in Table \ref{table:6}. This allows for a clearer understanding of the effectiveness of different prompts. The mean word accuracy of the Baseline model is 87.54\% and by utilizing just 1 minute of adaptation data with (Pad Cat, P+C) prompts, the performance is improved to 88.53\%. With more adaptation data of 3 and 5 minutes lengths, we can achieve the best word accuracies of 89.45\% and 89.99\% respectively. By comparing the different prompt methods, the padding prompt achieves the best performance among the single prompt methods for the LRW-ID dataset. This is the different tendency with that the addition prompt shows the best performance for the GRID dataset. It is related to that the visual front-end of LRW-ID is much deeper and has large parameters so that the addition prompt cannot largely affect the final encoded visual feature. On the other hand, the padding prompts can largely affect the visual feature at all intermediate CNN layers, with the large receptive field of deep CNN. Moreover, we observe similar results with GRID, the prompt (Add Pad, A+P) does not improve the performance over the padding-only prompt (Pad), and the prompt (Add Pad Cat, A+P+C) shows similar performance as (Pad Cat, P+C). The results indicate that the addition and the padding prompts have a similar role, improving the visual feature representations, thus the combination of them does not improve the performance further. However, the combination of prompts for CNN with the concatenation prompt which is for improving the temporal feature representations can boost the performance.

From the experimental results on both GRID and LRW-ID, we highlight that regardless of the types of prompts, we can improve the performance of pre-trained VSR models on the unseen target speaker by just finetuning prompts with a small number of adaptation data (\ie, less than 5 minutes long). In addition, when the pre-trained CNN model has deep architecture, the padding prompt has more effectiveness than the addition prompt, while the addition prompt can achieve better performance for shallow CNN. Finally, the combination of prompts for CNN and for Transformer can further improve the performance.

%------------------------------------ Table 8
%#############################################
\begin{table}[t]
	\renewcommand{\arraystretch}{1.4}
	\renewcommand{\tabcolsep}{2.0mm}
    \centering
    \caption{Performance (WER) comparisons with previous methods on GRID}
    \resizebox{0.999\linewidth}{!}{
	\begin{tabular}{ccccc}
	\Xhline{3\arrayrulewidth}
    \textbf{Method} & \textbf{1min} & \textbf{3min} & \textbf{5min} & \textbf{Total Params} \\ \hline
	Baseline \cite{assael2016lipnet} & 12.04 & 12.04 & 12.04 & 3.58M (+ 0\%) \\ \hdashline
    Speaker-invariant \cite{meng2018sitASR} & 11.28 & 11.28 & 11.28 & 3.58M (+ 0\%) \\ 
	Speaker code \cite{abdel2013codeicassp} & 5.56 & 4.83 & 4.68 & 3.66M (+ 2.21\%) \\ \hdashline
    FineTune-C & 10.01 & 8.81 & 8.46 & 3.62M (+ 1.14\%) \\
    FineTune-B & 5.57 & \underline{4.28} & 3.95 & 13.10M (+ 265.9\%) \\
    FineTune-F & \textbf{5.07} & \textbf{4.04} & \textbf{3.60} & 14.32M (+ 300\%) \\ \hline
	\textbf{Proposed Method (A+C)} & \underline{5.25} & 4.33 & 4.00 & 3.68M (+ 2.89\%) \\ 
    \textbf{Proposed Method (A+P+C)} & 5.53 & 4.31 & \underline{3.80} & 3.75M (+ 4.63\%) \\ 
	\Xhline{3\arrayrulewidth}
	\end{tabular}  \label{table:8}
	}
\end{table}
%#############################################

%------------------------------------ Table 9
%#############################################
\begin{table}[t]
	\renewcommand{\arraystretch}{1.4}
	\renewcommand{\tabcolsep}{1.8mm}
    \centering
    \caption{Performance (ACC) comparisons with previous methods on LRW-ID}
    \resizebox{0.999\linewidth}{!}{
	\begin{tabular}{ccccc}
	\Xhline{3\arrayrulewidth}
    \textbf{Method} & \textbf{1min} & \textbf{3min} & \textbf{5min} & \textbf{Total Params} \\ \hline
	Baseline \cite{assael2016lipnet} & 87.54 & 87.54 & 87.54 & 34.56M (+ 0\%) \\ \hdashline
    Speaker-invariant \cite{meng2018sitASR} & 88.22 & 88.22 & 88.22 & 34.56M (+ 0\%) \\ 
	Speaker code \cite{abdel2013codeicassp} & 88.09 & 88.76 & 89.08 & 35.59M (+ 2.97\%) \\ \hdashline
    FineTune-C & 87.71 & 87.79 & 87.88 & 39.43M (+ 14.1\%) \\
    FineTune-B & 88.39 & 88.96 & 89.52 & 398.83M (+ 1054\%) \\
    FineTune-F & 88.41 & 89.15 & \underline{89.91} & 691.21M (+ 1900\%) \\ \hline
    \textbf{Proposed Method (C)} & 88.28 & 88.92 & 89.33 & 34.61M (+ 0.15\%) \\ 
    \textbf{Proposed Method (P+C)} & \underline{88.53} & \textbf{89.45} & \textbf{89.99} & 37.68M (+ 9.01\%) \\ 
    \textbf{Proposed Method (A+P+C)} & \textbf{88.55} & \underline{89.39} & 89.75 & 37.93M (+ 9.74\%) \\ 
	\Xhline{3\arrayrulewidth}
	\end{tabular}  \label{table:9}
	}
\end{table}
%#############################################

\subsection{Comparison with previous methods}
In this experiment, we compare the effectiveness of the proposed prompt tuning methods with previous speaker-invariant and -adaptive methods including different finetuning methods. Especially, we also compare the total number of parameters required for inferring for all target speakers. For example, the method finetuning of the whole pre-trained model (\ie, FineTune-F) requires the parameters of the pre-trained model multiplied by the number of target speakers.
The comparison results on GRID dataset are shown in Table \ref{table:8}. We only represent the best two prompt types (Add Cat, A+C) and (Add Pad Cat, A+P+C) among all combinations, on the table. The best performance is represented in bold and the second-best performance is underlined. Among the methods that require less than 3\% additional parameters compared to that of the pre-trained model, Speaker-invariant, Speaker code, FineTune-C, and the proposed method (A+C), the proposed prompt tuning achieves the best performance on all adaptation data ranges. This shows the effectiveness of the proposed prompt tuning with a small number of additional parameters on speaker adaptation. Moreover, it is notable that compared to the method, Speaker code \cite{abdel2013codeicassp}, the proposed method has simpler adaptation stages by directly finetuning the prompts by attaching them to the pre-trained model. On the other hand, the method, Speaker code, has more adaptation stages where Adaptation Network should be firstly trained after attaching it to the pre-trained model on the training dataset, and then the speaker code should be finetuned on the adaptation dataset. Compared with the other finetuning methods, FineTune-B and FineTune-F, which consume more than 250\% additional parameters, the proposed prompt tuning methods, (A+C) and (A+P+C), show comparable performance with much fewer additional parameters (\ie, less than 5\% additional parameters). The results show that the proposed method is practical and is applicable to the speaker adaptation of tens of thousands of users in the real world while finetuning methods require astronomical memory for the speaker-specific models.

The comparison results on LRW-ID dataset are shown in Table \ref{table:9}. We only represent the two best prompt types, (Pad Cat, P+C) and (Add Pad Cat, A+P+C), and the smallest prompt type, (Cat, C), in the table. Surprisingly, in the adaptation setting of using under 5 minutes of adaptation data, the proposed methods even outperform the finetuning methods on LRW-ID. This result is also related to the previous studies \cite{jia2022vpt,lester2021prompttuning} that when the pre-trained model size becomes larger, the effectiveness of the prompt is getting bigger and even outperforms the finetuning methods. Please note that the model used for LRW-ID is about 10 times larger than that for GRID. Moreover, compared to the methods that require less than 15\% additional parameters for 20 target speakers, Speaker-invariant, Speaker code, FineTune-C, and the proposed methods, the proposed methods achieve the best performances. It is notable that using the concatenation prompt which increases just 0.15\% of model parameters outperforms the previous speaker-adaptive method, Speaker code.

%------------------------------------ Figure 3
%#############################################
\begin{figure}[t!]
	\centering
	\centerline{\includegraphics[width=9cm]{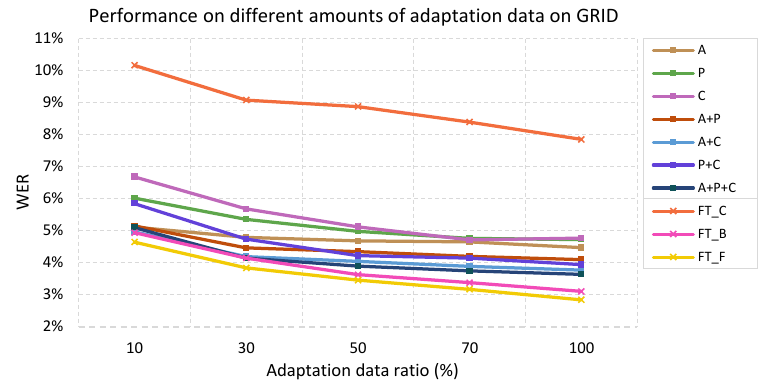}}
	\caption{WER comparisons between proposed prompt tuning and different finetuning methods according to adaptation data ratio on GRID.
	}
	\label{fig:3}
\end{figure}
%#############################################
%------------------------------------ Figure 4
%#############################################
\begin{figure}[t!]
	\centering
	\centerline{\includegraphics[width=9cm]{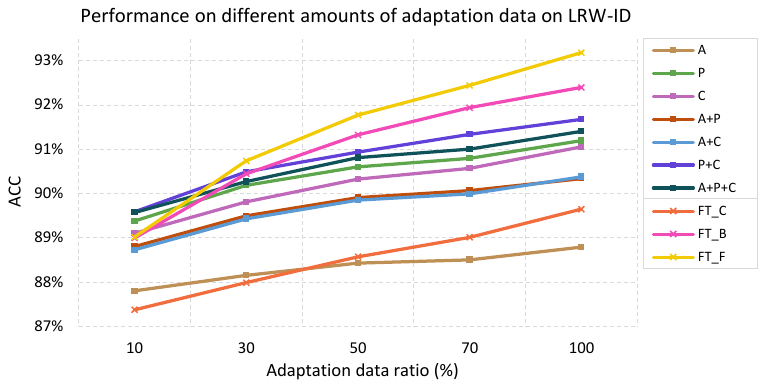}}
	\caption{ACC comparisons between proposed prompt tuning and different finetuning methods according to adaptation data ratio on LRW-ID.
	}
	\label{fig:4}
\end{figure}
%#############################################

\subsection{Adaptation results according to the data size}
In this experiment, we explore the performances of the proposed method without limiting the adaptation data size. Therefore, we analyze when the proposed prompt tuning is preferred over the finetuning methods. To this end, we utilize the different amounts of adaptation data proportionally to the entire adaptation set, 10\%, 30\%, 50\%, 70\%, and 100\%. Fig. \ref{fig:3} shows the adaptation results of the proposed prompt tuning and the three different finetuning methods on GRID. In overall, FineTune-B (FT-B) and FineTune-F (FT-F) outperform the other methods. Especially, the performance gap between FT-B (and FT-F) and the proposed prompt tuning gets bigger when more than 50\% adaptation data is utilized. The performance of the proposed prompt tuning (Add Pad Cat, A+P+C) with 100\% adaptation data is 3.13\% WER and the performance of the full finetuning, FineTune-F, is 2.34\% WER. Therefore, when the pre-trained model size is small, a better method can be selected by considering the model parameters and performance trade-off, based on the fact that the finetuning methods perform better while the proposed prompt tuning methods have much fewer parameters (\ie, 14.32M vs. 3.75M).

Fig. \ref{fig:4} shows the adaptation results on LRW-ID dataset. Compared to the finetuning methods, the proposed prompt tuning methods, (Add Pad Cat, A+P+C), (Pad Cat, P+C), (Pad, P), (Cat, C), show better results when the adaptation data is small so below 30\%. However, the finetuning methods, FineTune-B and FineTune-F outperform the prompt tuning methods when the adaptation data size becomes larger, and the performance gaps become larger according to the data size. Therefore, when the pre-trained model is large and there exist a few numbers of adaptation data, the prompt tuning method is preferred over the finetuning methods in the aspects of both performance and parameter size. When large-scale adaptation data is available for the target speaker, the trade-off between performance and additional parameters would be considered to select the appropriate speaker-adaptive method. Please note that the (A+P+C) prompt has 0.487\% parameters compared to the full model, for each speaker.

%------------------------------------ Table 10
%#############################################
\begin{table}[t]
    \renewcommand{\arraystretch}{1.5}
    \renewcommand{\tabcolsep}{2.2mm}
\centering
\caption{Performance (ACC) comparisons with adapters and LoRA on LRW-ID}
\resizebox{0.999\linewidth}{!}{
    \begin{tabular}{ccccc}
    \Xhline{3\arrayrulewidth}
    \textbf{Method} & \textbf{1min} & \textbf{3min} & \textbf{5min} & \textbf{Total Params} \\ \hline
    Baseline \cite{assael2016lipnet} & 87.54 & 87.54 & 87.54 & 34.56M (+ 0\%) \\ \hdashline
    Adapters (Att) \cite{he2022towards} & \textbf{88.93} & 89.42 & 89.82 & 35.61M (+ 3.03\%) \\ 
    Adapters (FFN) \cite{he2022towards} & 88.72 & 89.23 & 89.63 & 35.61M (+ 3.03\%) \\ 
    Adapters (Att+FFN) \cite{he2022towards} & 88.70 & 89.28 & 89.54 & 36.65M (+ 6.05\%) \\
    LoRA (rank 8) \cite{hu2022lora} & 88.51 & 88.98 & 89.44 & 36.53M (+ 5.69\%) \\
    LoRA (rank 16) \cite{hu2022lora} & 88.40 & 89.07 & 89.53 & 38.49M (+ 11.4\%) \\ \hdashline
    \textbf{Proposed Method (P+C)} & 88.53 & \textbf{89.45} & \textbf{89.99} & 37.68M (+ 9.01\%) \\ \hline
    \makecell{\textbf{Proposed Method (P+C)}\\ + Adapters (Att)} & 89.14 & 90.31 & 90.86 & 38.72M (+ 12.0\%) \\  
    \makecell{\textbf{Proposed Method (P+C)}\\ + LoRA (rank 16)} & 89.19 & 90.24 & 90.73 & 41.61M (+ 20.4\%) \\  
\Xhline{3\arrayrulewidth}
\end{tabular}  \label{table:param}
}
\end{table}

%#############################################

%------------------------------------ Table 11
%#############################################
\begin{table}[t]
	\renewcommand{\arraystretch}{1.2}
	\renewcommand{\tabcolsep}{4mm}
    \centering
    \caption{Ablation results (ACC) according to different padding prompt layers on LRW-ID}
    \resizebox{0.8\linewidth}{!}{
	\begin{tabular}{cccc}
	\Xhline{3\arrayrulewidth}
    \textbf{Adapt. min} & \textbf{$N_l=5$} & \textbf{$N_l=11$} & \textbf{$N_l=17$} \\ \hline
    1 & 88.09 & 88.38 & \textbf{88.42} \\
    3 & 88.28 & 88.70 & \textbf{89.32} \\
    5 & 88.55 & 88.99 & \textbf{89.62} \\
    \Xhline{3\arrayrulewidth}
	\end{tabular}  \label{table:10}
	}
\end{table}
%#############################################

%------------------------------------ Table 12
%#############################################
\begin{table}[t]
	\renewcommand{\arraystretch}{1.2}
	\renewcommand{\tabcolsep}{4mm}
    \centering
    \caption{Ablation results (ACC) according to different concatenation prompt lengths on LRW-ID}
    \resizebox{0.8\linewidth}{!}{
	\begin{tabular}{cccc}
	\Xhline{3\arrayrulewidth}
    \textbf{Adapt. min} & \textbf{$N_p=1$} & \textbf{$N_p=3$} & \textbf{$N_p=5$} \\ \hline
    1 & 88.27 & 88.14 & \textbf{88.28} \\
    3 & 88.61 & 88.68 & \textbf{88.92} \\
    5 & 89.08 & \textbf{89.41} & 89.33 \\
    \Xhline{3\arrayrulewidth}
	\end{tabular}  \label{table:11}
	}
\end{table}
%#############################################

\subsection{Comparison with parameter-efficient methods}
Recently, other parameter-efficient transfer learning methods \cite{he2022towards} such as adapters \cite{houlsby2019parameter} and Low-Rank Adaptation (LoRA) \cite{hu2022lora} have shown their effectiveness in finetuning Large Language Models (LLM) \cite{wu2023decoder,yeo2024visual,fathullah2024prompting} along with prompt tuning methods. In this experiment, we compare the proposed method with different types of these parameter-efficient transfer learning methods in speaker-adaptive VSR. To this end, 1) we insert trainable parallel adapters \cite{he2022towards}, each consisting of two linear layers with a reduction factor of 64, at either the attention layer, the feed-forward layer, or both layers of each Transformer; and 2) we fine-tune the trained models by applying LoRA to the query and key weights of each attention layer in the Transformer. We set both the rank and scale factor of LoRA to 8 or 16. Consequently, both the adapters and LoRA modules are trained on the adaptation set of each speaker while the pre-trained VSR model remains unchanged. Table \ref{table:param} shows the speaker adaptation results on LRW-ID. By comparing between adapters, LoRA, and the proposed prompt tuning methods, we can confirm that the proposed method outperforms the other methods in 3 minutes and 5 minutes settings, while the adapters at attention layer (\ie, Adapters (Att)) achieves the best result in the 1 minute adaptation setting. It is important to note that the proposed prompt tuning methods can be jointly applied with other parameter-efficient transfer learning methods. To verify this, we perform speaker adaptation by employing both the proposed prompt tuning and either adapters or LoRA. The last two rows in Table \ref{table:param} confirm that we can significantly improve performance when jointly applying either adapters or LoRA with the proposed prompt tuning methods. Considering both the number of parameters and performance, using the adapters at the attention layer combined with the proposed prompt tuning methods achieves the best performance compared to using with LoRA.

\subsection{Ablation study}
There are two hyperparameters that can be controlled, the number of CNN layers to insert the padding prompt and the length of the concatenation prompt. In order to confirm the effects of hyperparameters, we evaluate the adaptation performances by differing the number of layers (\ie, $N_l\in\{5,11,17\}$) that the padding prompt inserted and by differing the length of the concatenation prompt (\ie, $N_p\in\{1,3,5\}$). For the padding prompt ablation study, we only utilize the padding prompt without combination with other prompts to focus on the effects of the number of padding layers. Similarly, we only utilize the concatenation prompt for checking the effects of its length. The ablation results according to different padding prompt layers are shown in Table \ref{table:10}. When all padding layers are changed with the padding prompts, we can achieve the best performances for the three adaptation data settings (\ie, 1, 3, and 5 minutes). Moreover, we can find that the performance gains by using more layers become larger when more adaptation data is utilized. Therefore, we could further reduce the parameters for the padding prompt by inserting it for only the part of CNN layers, if the available adaptation data is less than 3 minutes. From the experimental result, we utilize all layers of CNN to insert the padding prompts in other experiments, as it shows the best results.
The ablation results according to different concatenation prompt lengths are shown in Table \ref{table:11}. For the concatenation prompt, we find that the length of the prompt does not affect the performance largely. We set the length of the concatenation prompt to 5 for other experiments, as it shows better performances overall.

\section{Conclusion}
In this paper, we proposed prompt tuning methods of DNNs for developing speaker-adaptive VSR models. With different types of prompts, the addition form, the padding form, and the concatenation form, we can adapt a pre-trained model composed of both CNN and Transformer variant architectures to the target unseen speaker. The proposed prompt has much fewer parameters compared to the full model, so it is more practical in storing and handling data for many users. Through comprehensive experimental evaluations, we showed that the padding prompt can improve the visual representations of CNN when the CNN is deep while the addition prompt is effective for shallow CNN. Moreover, by using the combination of the addition prompt and the padding prompt with the concatenation prompt, we can further increase the performance of VSR on unseen speakers. Finally, we compared the proposed method with several finetuning methods in the aspects of both the performance and the number of parameters, and showed that the proposed prompt tuning can achieve comparable VSR performances with much fewer additional parameters.

\section*{Acknowledgments}
This work was supported by Institute of Information \& communications Technology Planning \& Evaluation (IITP) grant funded by the Korea government(MSIT) (No.2022-0-00124, Development of Artificial Intelligence Technology for Self-Improving Competency-Aware Learning Capabilities)

\bibliographystyle{IEEEtran}
\bibliography{IEEEabrv,egbib}

\end{document}